\documentclass[runningheads]{llncs}

\usepackage[T1]{fontenc}
\usepackage[utf8]{inputenc}
\usepackage{graphicx}
\usepackage{amsmath}
\usepackage{booktabs}
\newcommand{\hd}[1]{\begin{tabular}[b]{@{}l@{}}#1\end{tabular}}
\begin{document}

\title{Enhancing AI Interpretability through Localised
Architectures}


\author{Ian Seet\inst{1} \and
Jonas Bozenhard\inst{2} \and
Simon Osterman\inst{3,4,5}}

\authorrunning{I. Seet et al.}

\institute{University of Glasgow, Glasgow, United Kingdom \and
Hamburg University of Technology, Hamburg, Germany \and
Saarland University, Saarbr\"ucken, Germany \and
German Research Center for Artificial Intelligence (DFKI),
Saarbr\"ucken, Germany \and
Centre for European Research in Trusted AI (CERTAIN),
Saarbr\"ucken, Germany}

\maketitle

\begin{abstract}
Recent advances in generative AI, especially powerful Large Language Models (LLMs), raise concerns over the interpretability, safety and sustainability of these large and opaque AI models. The power of such architectures is derived not only from the scalability of deep neural networks (DNNs), but also massively parallel hardware such as GPU clusters. The diffuse nature of DNNs gives them great function-approximation capability when provided with sufficient training data but imposes a cost in interpretability and computational efficiency. Observing that localised machine learning (ML) models tend to be more interpretable and computationally efficient than DNNs on small datasets, we reason by analogy that similar advantages may apply to specific localised hardware ML architectures. We argue that localised architectures with lower bandwidth but higher expressivity per node have the potential to be fundamentally more interpretable than DNNs running on GPU clusters while remaining competitive for smaller datasets. We then survey the suitability of various hardware ML paradigms for implementing such localised architectures and evaluate their per-node expressivity, energy efficiency and practical maturity of the technology required.

\keywords{Mechanistic interpretability \and AI safety \and
Localised architectures \and Neuromorphic hardware.}
\end{abstract}

\section{Introduction}

Over the past three decades, artificial intelligence (AI) has shifted from interpretable, computationally efficient algorithms based on expert knowledge to highly parallelisable machine learning models that perform far better but are neither efficient nor interpretable.~\cite{haenlein2019brief}.

These advances have been made possible by massively parallel hardware architectures, ranging from repurposed existing commercially available hardware such as Graphics Processing Units (GPUs)~\cite{raina2009largescale,ciresan2010deep} to bespoke designs such as Tensor Processing Units (TPUs) and Field-Programmable Gate Arrays (FPGAs)~\cite{jouppi2017indatacenter,chen2020survey}. These architectures have allowed the training and execution of highly scalable and performant but algorithmically inefficient and poorly-interpretable machine learning algorithms ranging from convolutional neural networks used in image processing~\cite{anwar2018medical} and zero-knowledge reinforcement learning~\cite{silver2018general} to transformer architectures capable of protein folding prediction~\cite{jumper2021highly}, image generation and natural language processing~\cite{devlin2018bert}.

Although powerful, the primary disadvantages of such AI systems -- poor interpretability and algorithmic inefficiency -- arguably raise serious risk. In the short-term, these limitations increase the difficulty of mitigating known hazards such as the generation of biased outputs and factually incorrect but plausible-sounding hallucinations. These risks, in turn, adversely affect the trustworthiness of such models~\cite{manzini2024should}. Moreover, the opacity of these algorithms increases the difficulty of aligning AI to human values, as it becomes challenging to determine precisely which values a model has learned. In the longer term, powerful but opaque Artificial General Intelligences (AGIs) may, on some accounts, pose an existential threat to humanity due to the potentially poor controllability of opaque models. This underscores that the opacity of AI systems built on deep neural networks (DNNs) poses critical risks today and in the future~\cite{bereska2024mechanistic,ngo2025alignment,weidinger2021ethical,gunning2019xai}. Improving the transparency and interpretability of AI systems is a key step toward making AI more controllable, safer, and better aligned with ethical values.

In addition, the algorithmic inefficiency of these models increases the computational resources required to train and operate them, which in turn increases energy consumption and CO\textsubscript{2} output with consequent impacts on climate change. In particular, Large Language Models (LLMs) consume enormous amounts of electricity and generate correspondingly large amounts of waste heat that demands cooling~\cite{manzini2024should}.

The poor interpretability of DNN architectures is a direct consequence of their highly interconnected and distributed nature, which often results in many features participating in high-order interactions that are computationally difficult to disentangle into weakly-interacting components. The same interconnectivity also increases the cost of training. Every connection between neurons carries its own weight parameter, so a densely connected network has a very large number of parameters, and each backward pass must compute a partial derivative with respect to every one of them.  

\section{Interpretability in machine learning and artificial intelligence}

Interpretability in machine learning (ML) has been defined~\cite{miller2019explanation,biran2017explanation} as ``the degree to which a human can understand the cause of a decision''. This framing is useful for policy and compliance contexts, where the primary concern is whether a human observer finds an explanation satisfying. However, such \emph{post hoc} explanations that satisfy a human observer do not guarantee that the explanation accurately reflects the model\textquotesingle s internal mechanisms, and a model that produces plausible-sounding explanations for wrong reasons is not thereby made safer or more controllable. A different and more demanding conception of interpretability is therefore required for AI safety: \emph{mechanistic interpretability}.

Mechanistic interpretability, as defined by Bereska et al.~\cite{bereska2024mechanistic}, involves ``reverse engineering the computational mechanisms and representations learned by an ML model into human-understandable algorithms and concepts to provide a granular, causal understanding.'' For a model to be mechanistically interpretable it is not sufficient to explain its outputs after the fact; the causal pathway from input to output must be understood and quantified at the level of the model\textquotesingle s internal components. This goal has been given formal grounding in recent theoretical work: Geiger et al.~\cite{geiger2025causal} define a mechanistic interpretation as a high-level causal model that is a faithful abstraction of a low-level model, where faithfulness is tested by performing interchange interventions and verifying that the high-level model correctly predicts the effects of those interventions on the low-level system. This framework makes mechanistic interpretability empirically falsifiable rather than merely descriptive.

In the philosophy of science, mechanistic interpretability has been compared to approaches used in the life sciences to investigate biological organisms~\cite{kastner2024explaining}. These approaches involve \emph{decomposition}, where phenomena are broken down into constitutive sub-functions (sometimes recursively to yield multi-level functional characterisations), and \emph{localisation}, which maps each sub-function to a specific part of the system thought to implement it~\cite{bechtel2010discovering}. Accordingly, the more extensively an AI system supports localisation, the greater its mechanistic interpretability. It is also useful here to note that mechanistic interpretability is model-specific while \emph{post hoc} interpretability tends to be more model-agnostic: mechanistic interpretability techniques, being dependent on model architecture, are not generalisable between different model types. Therefore, imposing design constraints (such as localisation) on model architecture is important if maximising mechanistic interpretability without compromising performance is desired.

A central obstacle to localisation in modern DNNs is the phenomenon of \emph{superposition}~\cite{elhage2022toy}. Under training pressure, neural networks are incentivised to represent more features than they have dedicated neurons, causing individual neurons to participate in the encoding of multiple unrelated concepts simultaneously. This \emph{polysemanticity} means that no single neuron can be assigned a clean functional role, and any attempt to localise a concept to a specific part of the network will typically fail: the concept is distributed across overlapping subsets of neurons in a way that is difficult to disentangle. Superposition is not an accidental property of particular architectures but an expected consequence of training highly connected networks with many more representational demands than available parameters. Any approach to interpretability that does not address superposition at the architectural level will face this as a persistent obstacle.

\section{Interpretability in neural networks vs. GAMs and tree-based models}

Although DNNs are generally the best-performing ML models in domains where data is plentiful, other models are more suitable when there is less data available, particularly if the inductive biases of DNNs are poorly aligned with the task. These alternative models are also often much more interpretable; two such classes of non-neural network models in particular serve as useful analogies when considering interpretability from a hardware perspective.

The first class of models we shall consider are Generalised Additive Models (GAMs)~\cite{wood2025generalized}. These models represent the output as a linear combination of smooth univariate functions \emph{f} of each of the \emph{n} features in the embedding.

\begin{equation}
y = \beta_{0} + f_{1}\left( x_{1} \right) + f_{2}\left( x_{2} \right) + \ldots + f_{n}\left( x_{n} \right)
\end{equation}

Since linear combinations are by definition additive, feature contributions in a basic GAM are very easy to isolate, giving GAMs a good reputation for both mechanistic and \emph{post hoc} interpretability. However, most real-world multivariable functions cannot be represented as a simple linear combination of univariate functions of each feature, and for better performance GAMs must include higher-order interaction terms \emph{g}.

\begin{equation}
y = \beta_{0} + f_{1}\left( x_{1} \right) + f_{2}\left( x_{2} \right) + \ldots + f_{p}\left( x_{p} \right) + g_{1}\left( x_{1},x_{2} \right) + g_{2}\left( x_{1},x_{3} \right) + \ldots
\end{equation}

These higher-order interaction terms become increasingly difficult to calculate as the number of these terms scales exponentially with the number of features. Thus, the interpretability of GAMs comes at the cost of performance for datasets dominated by high-order interactions. Nevertheless, GAMs remain competitive with neural networks on tabular data~\cite{doohan2026comparison}, particularly for small datasets. Tabular data is arranged like a spreadsheet: each row is a single observation and each column a distinct variable, typically with its own type and scale. Unlike the pixels of an image or the words of a sentence, the columns have no intrinsic ordering or adjacency — any two may be exchanged without loss of information. Consequently there is no spatial or sequential structure among features which can be exploited, which is why models that treat each feature separately remain competitive here.

The second class of models we shall consider are decision trees. The basic decision tree model models the output via a tree where each node other than the leaves represents a decision made based on only one feature; a rule is applied based on either a threshold (for continuous features) or a categorization (for discrete features)~\cite{blockeel2023decision}. The leaves of the tree represent the final outcomes. Basic decision trees are highly mechanistically interpretable since one can trace every decision made from root to leaf with clear thresholds or categories; however, they are prone to overfitting and become increasingly large and cumbersome for problems with a high feature dimension.

The two main approaches to mitigating these weaknesses are gradient boosting~\cite{chen2016xgboost} (e.g. XGBoost) and aggregation (e.g. Random Forest)~\cite{altman2017ensemble}. In the former, new trees intended to correct the errors of previous generations of trees are trained sequentially and the final prediction is the sum of the output of all the trees. In the latter, multiple trees are each trained on a subset of data and the final prediction is aggregated either by averaging (for continuous variables) or majority vote (for discrete variables). These approaches reduce interpretability but greatly improve the performance of the model. Indeed, for tabular data gradient-boosted decision trees generally outperform DNNs~\cite{liu2022performance,borisov2022deep}.

Although not considered as interpretable as standard decision trees, the fact that augmented decision trees do not differ significantly in their underlying mechanisms from standard decision trees illustrates how mechanistic interpretability contributes to \emph{post hoc} interpretability. SHAP (Shapley Additive exPlanations)~\cite{lundberg2017unified} is a \emph{post hoc} interpretability technique commonly used to attribute each feature's contribution to the model's prediction. Via the algorithm TreeSHAP, SHAP values can be efficiently calculated exactly not only for both basic decision trees, but even augmented decision tree models like XGBoost or Random Forest~\cite{yang2021fast}. This is because node splitting in augmented decision trees are still simple threshold functions based on a single feature, and this absence of high-order entanglement between features allows TreeSHAP to still efficiently attribute each feature contribution. In contrast, exact SHAP values cannot be efficiently calculated for DNNs and must be approximated~\cite{fernando2019study}.

Even though augmented decision trees are both more interpretable and perform better than DNNs for tabular data, node splitting has limited expressiveness in both basic and augmented decision trees (a split at a node is always a discrete function over a single feature rather than a higher-order function over multiple features). This limitation prevents decision trees from outperforming neural networks in problems with very large datasets and poorly defined features (e.g. image recognition and natural language processing).

\section{The Best of Both Worlds?}

A reasonable question to ask at this point is whether using a neural network to perform the node splitting in a decision tree leads to a model with both the interpretability of decision trees and the performance of neural networks on large non-tabular datasets. This idea has already been tested; certain neural tree models can outperform gradient-boosted decision trees for tabular data, albeit at the cost of interpretability~\cite{li2025survey}. GAMs can also be augmented with neural networks; Neural Additive Models (NAMs) replace each function \emph{f} in a GAM with a neural network operating on a single feature, giving greater expressivity than GAMs while retaining interpretability because the feature contributions remain additively separable~\cite{agarwal2021neural}. However, they retain the inability of GAMs to model higher-order cross-feature interactions.

At this point, it is worth thinking about how one might achieve maximum performance for a given interpretability requirement if computational resources were unlimited. One could imagine a modified decision tree where a large selection of discrete and continuous functions of varying complexity and order could be used during the node splitting instead of a simple threshold based on a single feature. In order to maintain interpretability and prevent overfitting, a cost is imposed on every non-linear interaction term proportional to its order during training so that such terms are only used when the increase in predictive power offsets the interpretability cost. Such a model would be much more expressive than a standard decision tree while remaining within a specified degree of interpretability due to the cost imposed on higher-order terms. However, it would also be extremely inefficient to train due to the exponential increase in search space for each function in the node-splitting library. It is therefore unlikely that such a model would be able to compete with DNNs for most practical problems, at least under the constraints of current hardware.

\section{How hardware influences ML models}

In the early era of computing, increases in processing power were largely achieved by increasing the clock speed of the processors. The power consumption of an electronic logic gate increases with at least the cube of clock speed~\cite{beloglazov2011chapter} and therefore simply overclocking a CPU to increase processing speed will result in large amounts of waste heat generated, which will eventually damage the device. Miniaturising circuit components reduces the capacitance and thus the energy required per clock cycle; however, quantum effects greatly increase noise as transistors shrink~\cite{equbal2023scaling}. This fundamental physical limit forced hardware architects to increase the parallelisation within their chip designs instead to continue making gains in processing power~\cite{millett2011future}, a trend exemplified by the massively parallel GPU and GPU-like chip architectures (e.g. TPUs and FPGAs) currently in widespread use.

This shift toward massively parallel chip architectures has also influenced the adoption of machine learning models that scale efficiently under parallel execution~\cite{mittal2019survey}. Access to massively parallel computing allows models to be trained directly on very high-dimensional representations, reducing the need for sophisticated feature engineering and causing in part the loss of interpretability which plagues deep learning. In the absence of massively parallel chip architectures, decision trees would likely see far more widespread use as they do not benefit from GPU parallelisation to the same extent that DNNs do~\cite{wen2018efficient}.

\section{Locality, interpretability and distributed computing}

Throughout this paper, we have tried to emphasize the overarching theme that interpretable ML models tend to be localised -- that is, they do their best to avoid entangling many features together in high-order interactions unless doing so substantially improves the predictive power of the model. It is important here to observe a critical nuance: localised algorithms can still benefit greatly from parallelisation; for instance, a parallel processor could calculate each term in a GAM separately before the final summation which is the only step not trivially parallelisable.

The key difference, however, is that GPU clusters are not merely parallel but distributed; large quantities of information can be transmitted between GPUs and memory because electronic information transmission has both a high bandwidth and low latency compared to signal transmission through the axon of a biological neuron~\cite{testasilva2014high,anon2018nvidia}, for instance. The fact that information transmission in GPUs has a high bandwidth and low latency is critical for backpropagation in DNNs because dense connectivity between layers requires large volumes of gradient information to be exchanged and synchronised efficiently during training.

If the available hardware were not capable of this high-bandwidth-low-latency information exchange, it would become necessary to design more localised (and thus more interpretable) ML models. To compensate for this reduction, the architecture would need to be more expressive at each individual node. A precedent for this type of hardware architecture functioning in practice exists in the form of biological neural networks, which as previously stated have far lower bandwidth and higher latency than electronic logic.

It would be difficult to describe the operation of biological neural networks as interpretable in the same sense applied to artificial ML models. However, this difficulty arises less from an inherent opacity of localised, low-bandwidth neural architectures than from a mismatch between the functions expressible by biological neurons (e.g. in dendritic computations~\cite{acharya2022dendritic}) and the analytic and discrete functions that form the basis of human mathematical description. Indeed, the localised nature of biological neural networks allows the portions of the human brain to be targeted directly to alter a person's behaviour for medical purposes, e.g. with deep brain stimulation~\cite{perlmutter2006deep}.

Furthermore, the localised nature of biological neural networks has inspired artificial analogues. By deliberately imposing a constraint to minimise the connection cost between neurons, Liu et al. report that Brain-Inspired Modular Training (BIMT) increases modularity and interpretability with only slight performance degradation \cite{liu2024seeing}, albeit only on small-scale tasks. However, as neuron positions are arbitrarily defined for bookkeeping purposes rather than physical locations, locality here is obtained through regularisation and not intrinsic to the substrate as is the case for biological networks.

It is also useful here to note a difference between interpretability and \emph{controllability}. A model does not have to be interpretable in either the mechanistic or \emph{post hoc} sense for an external control to modify its output in a predictable manner, and it should be intuitively obvious that this is easier to accomplish for localised models than models where representations are highly distributed like artificial DNNs -- hence, for instance, the practice of using Reinforcement Learning with Human Feedback (RLHF) to alter the behaviour of LLM transformers over trying to identify and modify specific parts of the model responsible for undesirable behaviour~\cite{chaudhari2025rlhf}. Controllability, in this sense, refers to the ability to induce reliable and targeted changes in model behaviour through external intervention, without necessarily understanding the underlying computational mechanisms that produce that behaviour. It is a weaker but more practically attainable property than full mechanistic interpretability, and one that is increasingly important as AI systems grow more powerful and broadly deployed.

A prominent example of controllability without full mechanistic interpretability is activation steering, wherein targeted additive interventions on intermediate model representations are used to shift model behaviour at inference time, without any parameter updates~\cite{turner2024activation,ostermann2026weights,zou2023representation}. Such methods are grounded in the Linear Representation Hypothesis (LRH), which posits that concepts tend to be encoded as approximately linear directions in a model\textquotesingle s activation space~\cite{park2023linear}. Crucially, reliable controllability does not require the LRH to hold universally: Recent work has shown that when linearity assumptions are relaxed, some representational alignment claims become vacuous~\cite{sutter2026nonlinear}. It is sufficient that approximate linear structure is stable enough to support predictable intervention in practice; even if the underlying geometry is non-linear, a linear approximation over local regions of activation space yields consistent behavioural effects across a diverse range of models and tasks~\cite{arditi2024refusal,rimsky2024steering}. Where such structure is absent or unstable, steering may fail, which is a limitation that itself carries diagnostic value for understanding model internals. While full mechanistic interpretability would be ideal, controllability in its absence can still improve safety by allowing for targeted modifications to curb undesirable behaviour.

\section{Implementing localised ML hardware architectures}

In general, a localised ML hardware architecture should have two characteristics if it is to be competitive against distributed architectures while maintaining interpretability. Firstly, the high latency and low bandwidth must be compensated for to render these architectures competitive against their distributed counterparts. This can be done by increasing the expressivity of each local node, although there must be a cost imposed on higher-order interactions to maintain interpretability per the example of the modified decision tree.

Secondly, high energy efficiency is also desired. It would be naive to expect the AI industry to switch to localised architectures for interpretability alone while sacrificing some performance as such a switch would likely not be profitable. If, however, the localised architectures were more energy efficient, it would be possible to encourage private businesses to make the switch through financial incentives, for instance via the imposition of taxes on energy-hungry data centres.

The seemingly obvious choice to implement such an architecture are analogue electronics~\cite{belostotski2025survey}, which have already been used to both run and train neural networks. As one major selling point of analogue electronic architecture is its enhanced energy efficiency compared to digital architectures when running neural networks, it would seem to be a logical choice given the criteria outlined above. However, several major problems would limit the effective implementation of localised architectures with analogue electronics.

Firstly, analogue logic is generally susceptible to errors induced by thermal noise, and while analogue electronics are often far more efficient than digital electronics when running neural networks, they are often less efficient during training because the higher precision required when calculating gradients force costly analogue-digital conversions and error correction steps~\cite{fahimi2021mitigating,rasch2024fast}. Perhaps the larger issue, however, is that electronic analogue logic is simply not as expressive per unit bandwidth when compared to other architectures (for instance, electrochemical systems like biological neural networks~\cite{beniaguev2021single}); thus, there is little reason for electronic analogue ML models to differ significantly in localisation from models designed to run on GPUs. In addition, the noise sensitivity of analogue systems complicate the identification of precise states or transitions underlying particular system behaviours and outcomes, thereby negatively affecting interpretability in practice and in real-life applications~\cite{kiener2026ethics}.

If electronic analogue logic is insufficiently expressive, we could instead consider electrochemical neuromorphic architectures, which are inspired by biological neural networks and exploit charge separation and ion concentration gradients to encode, process and transmit information~\cite{chen2023electrochemicalmemristorbased}. While there is a clear precedent for the broad concept working in practice (the human brain), we would likely encounter similar problems with interpretability that also occur when analysing the behaviour of biological neural networks. It could be argued that unlike biological neurons, an artificial electrochemical neuromorphic architecture could be designed from the ground up to use only activation functions considered interpretable while still being more expressive than artificial neural networks. However, modelling electrochemistry and ion dynamics at the nanoscale (the same scale biological voltage-gated ion channels operate at) with the accuracy needed to implement interpretable well-defined nonlinear activation functions would be highly computationally demanding~\cite{flood2019atomistic,modi2012computational}. Conversely, it is less likely that electrochemical devices operating on larger scales (where their behaviour would be easier to model with bulk diffusion processes) would be fast or efficient enough to compete against neural networks running on traditional distributed electronic computing architectures.

If electrochemical systems are too difficult to model at the scales necessary to be competitive with traditional electronic computing, we could in turn consider purely chemical architectures. Chemical reaction networks have been used for this purpose~\cite{baltussen2024chemical}. However, many of the same arguments applied to electrochemical systems apply to diffusion-based chemical architectures; the reduction in scale required to compete against traditional electronics would also make the task of modelling the architecture to the accuracy required computationally difficult. In addition, long-range transmission of information in a pure chemical diffusion system would be even slower than an electrochemical system, which is already orders of magnitude slower than traditional electronics. There is irrefutable physical evidence in the form of the human brain that information transmission at the speeds possible in an electrochemical system still allow low-bandwidth electrochemical ML architectures capable of outperforming electronic ML architectures across many domains to be designed, but no corresponding evidence that the even slower transmission speed in pure chemical diffusion systems would allow for similarly viable chemical architectures.

If our objective was purely to create an analogue hardware architecture which could implement a selected function to the highest accuracy possible, macroscale mechanical logic (e.g. the Babbage engine) would be the most reasonable choice given that these systems can be modelled to a very high accuracy with only pure Newtonian mechanics~\cite{yasuda2021mechanical}. Macroscale mechanical logic is unfortunately both slow and inefficient by the standards of electronic logic and would not be a viable competitor against the latter.

To render mechanical logic viable, it would likely be necessary to miniaturise the devices to the point where thermal noise becomes significant~\cite{seet2023simulation}. These molecular mechanical devices execute logical operations and transmit information via conformational changes of highly rigid molecules and are of sufficiently low mass such that the average kinetic energy of the system, while in operation, is not significantly greater than the kinetic energy of the system when in thermodynamic equilibrium. By avoiding bond formation and diffusion, molecular mechanical systems are both highly energy efficient and can be accurately simulated with purely Newtonian molecular dynamics~\cite{seet2023simulation,seet2023pipelined}, far less computationally intensive than the fluid dynamics needed for accurate modelling of electrochemical and chemical diffusion-based systems.

However, since these systems are subject to the effects of thermal noise, analogue-to-digital conversion for error correction and information transmission would also be as necessary as it is for analogue electronic logic. Even so, thermal noise is not necessarily a pure detriment; thermal noise in a molecular system could also potentially be exploited to improve training by using an annealing-based algorithm~\cite{lee2012hardware} to find good approximate solutions to the difficult combinatorial problem of minimising the number of few high-order interactions while maximising predictive power. Molecular mechanical systems should also possess a high degree of expressiveness due to having many more degrees of freedom compared to an electronic analogue circuit, though it is likely there will be functions the latter can approximate more efficiently. These advantages in efficiency and expressiveness are offset by the fact that such systems are currently not practical to synthesise via either bottom-up chemical synthetic techniques or top-down lithography techniques.

\begin{table}[htbp]
\caption{Comparison of candidate hardware paradigms for implementing
localised machine learning architectures.}\label{tab:paradigms}
\begin{tabular}{@{}lccccl@{}}
\toprule
Paradigm & \hd{Expressivity\\per bandwidth} & Latency & Noise & \hd{Energy\\efficiency} & Maturity\\
\midrule
Digital electronic   & Low  & Very low  & Low      & Low      & Standard\\
Analogue electronic  & Low  & Very low  & Moderate & Moderate & Extant\\
Electrochemical      & High & Moderate  & High     & Moderate & Speculative\\
Chemical diffusion   & High & High      & High     & Low      & Speculative\\
Mechanical           & High & Very high & Very Low & Very low & Extant\\
Molecular mechanical & High & Low       & High     & High     & Speculative\\
\bottomrule
\end{tabular}
\end{table}

\section{Conclusion}

In this paper, we discuss the concept of interpretability and the value of mechanistic interpretability in particular for improving AI safety. By contrasting the fundamental structure of two classes of interpretable ML models with DNNs, we argue that localisation tends to make a model more interpretable, and the physical characteristics (high bandwidth and low latency) of distributed electronic hardware architectures specifically encourage the development of uninterpretable ML models.

Each hardware paradigm we have proposed to potentially implement localised architecture has distinct advantages and disadvantages. Electronic analogue logic is the most mature but lacks expressiveness. Electrochemical and chemical diffusion-based architecture most closely resembles biological neural networks but are slower and fundamentally difficult to simulate to high accuracy. Molecular mechanical logic is easy to simulate to high accuracy, highly expressive and potentially very energy efficient, but the furthest from practical synthesis. Despite the potential difficulties in implementing these localised interpretable architectures, we believe that the threats posed from uninterpretable AI justifies continued research in this domain.

\begin{credits}
\subsubsection{\ackname}
This research received no external funding. The authors are grateful to
Kevin Baum for reviewing the manuscript and providing feedback.

\subsubsection{\discintname}
The authors have no competing interests to declare that are relevant to
the content of this article.
\end{credits}

%
%
\bibliographystyle{splncs04}
\bibliography{references}

\end{document}